\documentclass[letterpaper, 10 pt, conference]{ieeeconf}

\IEEEoverridecommandlockouts
\usepackage{tabularx}
\usepackage{multirow}
\usepackage[table]{xcolor}

\newcolumntype{Y}{>{\centering\arraybackslash}X}

\usepackage{amsmath}
\usepackage{amssymb}
\usepackage{array}
\usepackage{cite}
\usepackage{graphicx}
\usepackage{booktabs}
\usepackage{url}

\newif\ifhasreferences
\hasreferencestrue

\title{\LARGE \bf
{Colon3R}: Cross-Domain 3D Reconstruction from Monocular Colonoscopic Video
}

\author{Zhihao Xing, Yingyu Wang, Liang Zhao, Shoudong Huang}

\begin{document}

\maketitle
\thispagestyle{empty}
\pagestyle{empty}

\begin{abstract}
    Monocular colonoscopic 3D reconstruction is important for surgical robotic
    colonoscopy, but remains challenging due to weak texture, specular reflections, limited
    view overlap, and non-rigid tissue motion. Conventional multi-view 3D reconstruction methods rely on stable correspondences and approximate
    rigidity, which are often violated in colonoscopy. Existing endoscopic methods often rely on domain-specific supervision, whereas
    there are not enough in-vivo labeled data available to adapt geometry foundation models to clinical colonoscopy. We present Colon3R, a cross-domain semi-supervised framework  built on pretrained VGGT that transfers
    coupled camera, depth, and pointmap geometry from labeled phantom and simulated
    data to unlabeled in-vivo colonoscopy without requiring target-domain geometric
    annotations. Unlike source-only fine-tuning, which learns only from phantom and simulated
    data, Colon3R directly exploits unlabeled in-vivo video through teacher-derived
    cross-view supervision. Our proposed hierarchical quasi-rigid
    reliability selects reliable supervision at the sequence, directed-pair, and
    pixel levels, while source-preserving adaptation retains the learned coupled
    geometry during target-domain adaptation. Extensive experiments demonstrate that our method achieves superior overall
    performance over state-of-the-art approaches in depth, pointmap, and
    camera pose estimation. Qualitative comparisons on real in-vivo colonoscopy
    further show substantially more complete and geometrically consistent
    reconstructions than competing methods under clinical domain shift. The code is publicly available at \url{https://github.com/vglx/Colon3R}.
\end{abstract}

\section{Introduction}

Colonoscopy is an important clinical procedure for examining and treating the
lower gastrointestinal tract~\cite{Tumino2023RoboticColonoscopy}, and provides
a natural setting for computer-assisted and robotic
intervention~\cite{Obstein2026MFE}. For surgical robotic systems, 3D
reconstruction from monocular colonoscopic video provides a spatial
representation of the observed environment, supporting endoscope localisation
and mapping~\cite{Ozyoruk2021EndoSLAM}, as well as downstream tasks such as
coverage assessment and registration
with preoperative anatomical models~\cite{Ma2021RNNSLAM}~\cite{Bobrow2023C3VD}.

However, monocular reconstruction in colonoscopy is challenging
because the endoscope operates at close range within a narrow and deformable
lumen, which can cause unstable tracking and accumulated geometric
drift~\cite{Ozyoruk2021EndoSLAM}~\cite{Ma2021RNNSLAM} over extended colonoscopic
sequences. Classical visual odometry, simultaneous localisation and mapping
(SLAM), and multi-view reconstruction typically rely on stable cross-view
correspondences and approximately rigid scene geometry to constrain camera
motion and 3D structure. Weak or repetitive texture, specular reflections, and
limited visibility make reliable correspondence estimation
difficult~\cite{Chu2020EndoscopicMatching}.

Learning-based methods improve endoscopic geometry estimation by
predicting depth and camera motion directly from images. Rau \textit{et al.}~\cite{Rau2019DomainAdaptation} explored domain adaptation for monocular
endoscopic depth estimation, while EndoSLAM~\cite{Ozyoruk2021EndoSLAM}
combines learned depth with visual odometry. More recent approaches such as
Surgical-DINO~\cite{Cui2024SurgicalDINO},
EndoDAC~\cite{Cui2024EndoDAC}, and
EndoDAV~\cite{zhou2025endodav} adapt pretrained visual representations for
endoscopic depth estimation and temporal prediction. However, these methods remain largely depth-centered, with coherent 3D
reconstruction relying on separately estimated camera motion or additional
geometric alignment, where inconsistencies between depth and pose can lead to
cross-view misalignment and reconstruction drift. Moreover, the domain gap
between controlled phantom or simulated training data and real in-vivo
colonoscopy cannot be fully addressed by source-only fine-tuning, motivating
adaptation with unlabeled target-domain observations~\cite{Jiang2025ColonAdapter}.

Recent feed-forward geometry foundation models provide a more general
formulation for multi-view reconstruction. Trained on diverse 3D data, they
learn transferable geometric priors that generalize beyond task-specific
depth or pose estimation. DUSt3R~\cite{Wang2024DUSt3R}
formulates pairwise reconstruction as dense pointmap regression, while
MASt3R~\cite{Leroy2024MASt3R} strengthens correspondence estimation.
Fast3R~\cite{Yang2025Fast3R} and MUSt3R~\cite{Cabon2025MUSt3R} further
extend this paradigm to multi-view inputs. VGGT~\cite{Wang2025VGGT} instead predicts camera poses, camera-axis depth,
dense pointmaps, and tracks for all input views in a single forward pass.
Its camera translations and depth maps are scale-consistent across the input
sequence, with pointmaps expressed in the first-camera reference frame. This enables cross-view constraints to be constructed directly from the
predictions, avoiding the additional pose recovery or global alignment required
by pairwise pointmap-based methods and reducing extra sources of geometric
estimation error during adaptation, which makes VGGT particularly suitable for
our cross-domain setting.

Their endoscopic extensions further demonstrate the potential of geometry
foundation models. Endo3R~\cite{Guo2026Endo3R} addresses online geometry
prediction in dynamic endoscopy, while ColonAdapter~\cite{Jiang2025ColonAdapter}
adapts pointmap-based foundation geometry to colonoscopy through pairwise
prediction and global alignment. However, these models do not specifically
address adaptation from geometrically annotated phantom and simulated data to
unlabeled in-vivo colonoscopy, where clinical appearance and tissue deformation
introduce substantial domain shift.

\begin{figure*}[!t]
    \centering
    \includegraphics[width=0.98\textwidth]{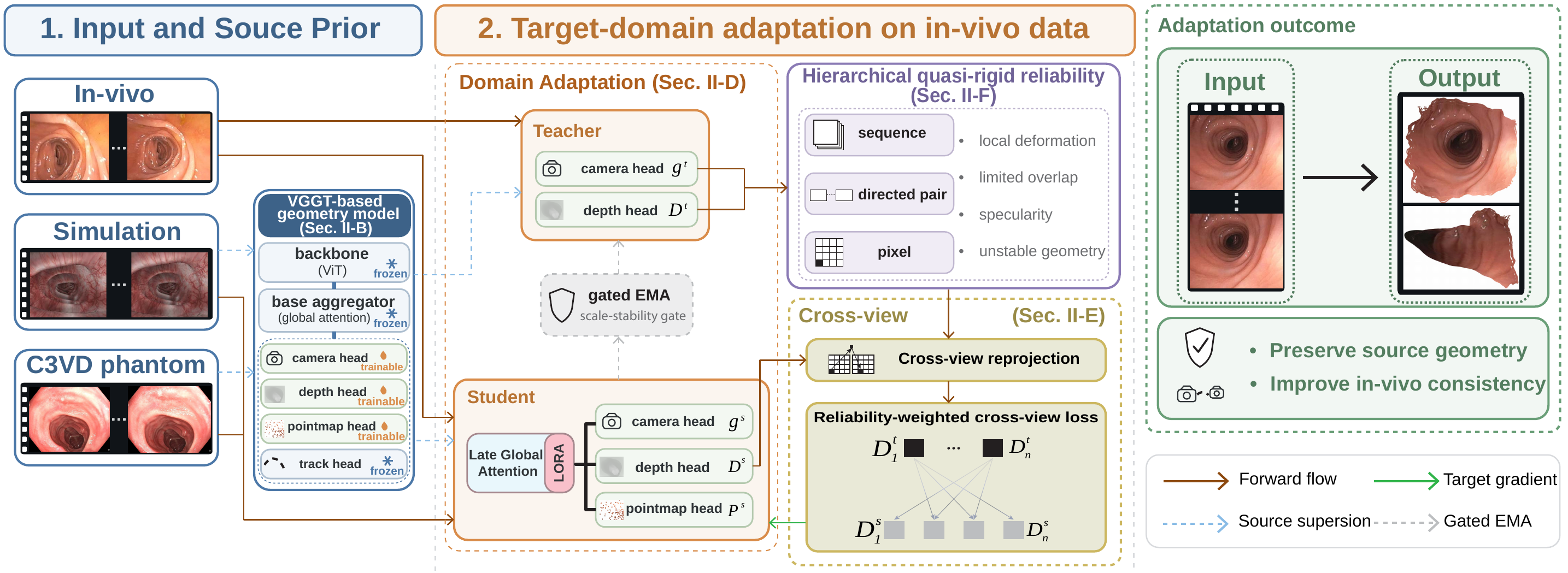}
    \caption{
    Overview of Colon3R. A source-supervised VGGT geometry prior is adapted to
    unlabeled in-vivo video through a parameter-efficient teacher--student
    framework with scale-stability-gated exponential moving average (EMA) and hierarchical quasi-rigid
    reliability-weighted cross-view supervision. The adaptation improves in-vivo
    geometric consistency while preserving the coupled source geometry.
    }
    \label{fig:colon3r_overview}
\end{figure*}

Direct geometric supervision on real in-vivo colonoscopy is difficult because
metric geometric ground truth is not readily available during clinical
procedures. Labeled phantom datasets~\cite{Bobrow2023C3VD}
and simulated colonoscopy data~\cite{zhang20213d} therefore provide reliable
geometric supervision. Yet these controlled data cannot fully reproduce the
clinical appearance and tissue deformation encountered in-vivo, creating an
asymmetric supervision gap between labeled source data and unlabeled real
observations.

Unlabeled in-vivo video can nevertheless provide cross-view geometric
constraints for target-domain adaptation~\cite{ren2026wat3r}~\cite{Jiang2025ColonAdapter}.
In colonoscopy, however, tissue deformation, limited overlap, and visibility
changes can make these constraints unreliable. Their reliability can vary
across temporal sequences, directed view pairs, and local image regions, so
uniformly applying them may introduce incorrect supervision. These observations motivate our reliability-aware adaptation strategy, which
selectively uses cross-view constraints according to their geometric
reliability.

Accordingly, we propose Colon3R, a cross-domain
semi-supervised framework for monocular colonoscopic 3D reconstruction.
Colon3R first initializes from pretrained VGGT and learns a coupled geometry
prior using labeled phantom and simulated data, which is then adapted to
unlabeled in-vivo video through teacher--student learning. During adaptation, hierarchical quasi-rigid reliability replaces uniform
cross-view supervision with selective weighting at the sequence, directed-pair,
and pixel levels, suppressing constraints that become unreliable under tissue
deformation, limited overlap, occlusion, or specular reflection. Source-preserving
updates further retain the geometry learned from labeled source data. This
enables reliable adaptation to unlabeled in-vivo video without geometric
annotations or explicit deformation estimation.
The main contributions of this paper are summarized as follows:
\begin{itemize}
    \item We propose Colon3R, a cross-domain semi-supervised framework that transfers
    coupled geometry from labeled phantom and simulated data to unlabeled in-vivo
    colonoscopy, improving geometric reconstruction under clinical domain shift.

    \item We propose hierarchical quasi-rigid reliability to identify reliable
    cross-view constraints at sequence, directed-pair, and pixel levels under
    tissue deformation, improving the robustness of target-domain supervision
    without explicitly estimating deformation.

    \item We introduce source-preserving adaptation that combines reliable target
    supervision with continued source supervision and constrained updates, enabling
    adaptation to in-vivo video while preserving the coupled geometry learned from
    labeled source data.

    \item Extensive experiments demonstrate superior overall performance over
    state-of-the-art approaches across different colonoscopic settings. In
    particular, results on real in-vivo colonoscopy show that our method produces significantly
    more complete and geometrically consistent reconstructions under clinical
    domain shift.
\end{itemize}
\section{Method}

Our method follows a two-stage cross-domain learning framework, as illustrated
in Fig.~\ref{fig:colon3r_overview}. First, labeled phantom and simulated data
are used to establish a source-supervised VGGT~\cite{Wang2025VGGT} geometry prior
(Sec.~\ref{subsec:source_prior}). The resulting prior is then adapted to unlabeled
in-vivo video through teacher--student learning
(Sec.~\ref{subsec:adaptation}). During adaptation, teacher-derived cross-view
supervision is constructed from unlabeled target video
(Sec.~\ref{subsec:crossview_supervision}), while hierarchical quasi-rigid
reliability evaluates its validity at the sequence, directed-pair, and pixel
levels (Sec.~\ref{subsec:hierarchical_reliability}).

\subsection{Problem Definition}
\label{subsec:problem}

We formulate Colon3R by adapting the standard pretrained
VGGT~\cite{Wang2025VGGT}, retaining its joint prediction of camera pose,
camera-axis depth, and direct pointmaps. Given a monocular colonoscopy image
sequence \(\mathcal{I}=\{I_i\}_{i=1}^{T}\), the resulting predictor
\(f_{\Theta}\) maps the input sequence to
\begin{equation}
    f_{\Theta}(\mathcal{I})
    =
    \left\{
        \mathbf{T}_{i},
        \mathbf{D}_{i},
        \mathbf{P}_{i}
    \right\}_{i=1}^{T},
    \label{eq:geometry_mapping}
\end{equation}
where \(\mathbf{T}_{i}\) is the world-to-camera transformation,
\(\mathbf{D}_{i}\in\mathbb{R}^{H\times W}\) is the depth map,
and \(\mathbf{P}_{i}\in\mathbb{R}^{H\times W\times 3}\) is the direct
pointmap expressed in the first-camera reference frame.

Training uses labeled phantom and simulated sequences with geometric
annotations together with unlabeled in-vivo RGB sequences. The objective is
to adapt \(f_{\Theta}\) to in-vivo colonoscopy using the reliable source
supervision and geometry from unlabeled target video,
without requiring in-vivo geometric ground truth, while retaining the coupled
geometry learned from the source domain.

\subsection{Source-Supervised Geometry Prior}
\label{subsec:source_prior}

To establish a geometry prior for subsequent adaptation to
unlabeled in-vivo colonoscopy, we first train the predictor \(f_{\Theta}\) in
Eq.~\eqref{eq:geometry_mapping} on the labeled phantom and simulated data. Starting from the pretrained
VGGT parameters, the three geometric outputs are supervised using the standard
VGGT objectives~\cite{Wang2025VGGT}:
\begin{equation}
    \mathcal{L}_{\mathrm{src}}
    =
    \mathcal{L}_{\mathrm{cam}}
    +
    \mathcal{L}_{\mathrm{depth}}
    +
    \mathcal{L}_{\mathrm{point}},
    \label{eq:source}
\end{equation}
where the three terms correspond to camera pose, camera-axis depth, and direct
pointmap supervision, respectively. Further details of these objectives can be
found in~\cite{Wang2025VGGT}.

The resulting source-supervised predictor \(f_{\Theta_{0}}\), where
\(\Theta_{0}\) denotes the parameters obtained after source-supervised
training, serves as the coupled geometry prior and initializes both the student
and teacher for subsequent adaptation to unlabeled in-vivo video. For
comparison, continuing the same source-only optimization from this checkpoint
without target-domain adaptation yields the Fine-tuned VGGT baseline
used in our experiments.

\subsection{Cross-Domain Training Objective}
\label{sec:optimization}

To adapt to the in-vivo domain without degrading the source-learned geometry,
we jointly optimize labeled source supervision and supervision from unlabeled
target video. The overall objective is
\begin{equation}
    \mathcal{L}_{k}
    =
    \frac{1}{M_s}
    \sum_{\nu=1}^{M_s}
    \mathcal{L}^{(\nu)}_{\mathrm{src}}
    +
    \lambda_t(k)
    \left(
        \mathcal{L}_{\mathrm{cv}}
        +
        \lambda_{\mathrm{anc}}\mathcal{L}_{\mathrm{anc}}
    \right),
    \label{eq:optimization}
\end{equation}
where \(k\) indexes the training updates,  \(M_s\) is the number of labeled source microbatches accumulated in each
student update, and \(\mathcal{L}^{(\nu)}_{\mathrm{src}}\) is the source-supervised geometry
objective defined in Sec.~\ref{subsec:source_prior}  evaluated on the
\(\nu\)-th source microbatch,
\(\mathcal{L}_{\mathrm{cv}}\) denotes the reliability-weighted cross-view
objective detailed in the following subsections, and
\(\mathcal{L}_{\mathrm{anc}}\) constrains the student depth scale relative to
the fixed source-supervised reference. The coefficients \(\lambda_t(k)\) and \(\lambda_{\mathrm{anc}}\) control the
target-domain supervision and depth-scale regularization, respectively. The following subsections describe how the target cross-view supervision is
constructed and how its reliability is evaluated under non-rigid colonoscopic
observations.

\subsection{Semi-Supervised Adaptation}
\label{subsec:adaptation}

Since geometric annotations are unavailable in the target domain, we adopt a
teacher--student framework~\cite{Tarvainen2017MeanTeacher}
to provide target-domain geometric guidance while adapting the source prior.
The student learns from both labeled source data and teacher-derived target
geometry, while the teacher provides detached predictions for the target
branch. Both networks are initialized as identical copies of the
source-supervised predictor \(f_{\Theta_0}\) obtained in
Sec.~\ref{subsec:source_prior}:
\begin{equation}
    \Theta^{\mathrm{stu}}_{0}
    =
    \Theta^{\mathrm{tea}}_{0}
    =
    \Theta_{0}.
    \label{eq:student_teacher_init}
\end{equation}
The student learns from both labeled source supervision and teacher-derived
target supervision, while the teacher predictions are detached from gradient
computation.

To preserve the source-learned coupled geometry during target adaptation,
source and target supervision update different subsets of the student
parameters:
\begin{equation}
\begin{aligned}
    \mathcal{P}_{\mathrm{src}}
    &=
    \{\phi,\theta_{\mathrm{cam}},
    \theta_{\mathrm{depth}},\theta_{\mathrm{point}}\},\\
    \mathcal{P}_{\mathrm{tgt}}
    &=
    \{\phi,\theta_{\mathrm{depth}}\}.
\end{aligned}
\label{eq:branch_updates}
\end{equation}
where \(\phi\) denotes the LoRA parameters~\cite{Hu2022LoRA}, and
\(\theta_{\mathrm{cam}}\), \(\theta_{\mathrm{depth}}\), and
\(\theta_{\mathrm{point}}\) denote the three geometry heads. The LoRA adapters
are inserted into selected late global-attention blocks, while the VGGT~\cite{Wang2025VGGT}
backbone, base aggregator, and track head remain frozen. This design allows
source supervision to constrain the coupled geometry while preventing
target-domain supervision from directly updating the camera and pointmap
heads.

Teacher-derived supervision can become unstable when rapid student changes are
propagated to the teacher. We therefore keep the teacher fixed during the
first \(K_{\mathrm{h}}\) student updates and synchronize it once with the
current student at the end of the warm-up:
\begin{equation}
    \Theta^{\mathrm{tea}}_{K_{\mathrm{h}}}
    =
    \Theta^{\mathrm{stu}}_{K_{\mathrm{h}}}.
    \label{eq:teacher_sync}
\end{equation}
Subsequent teacher updates are accepted only when the current student state
has sufficient geometric support and stable depth scale:
\begin{equation}
    q_k
    =
    \mathbb{I}\!\left[S_k\geq\tau_{\mathrm{sup}}\right]
    \mathbb{I}\!\left[
        \Delta_{\mathrm{scale}}^{(k)}
        \leq\tau_{\mathrm{scale}}
    \right],
    \qquad k>K_{\mathrm{h}},
    \label{eq:ema_gate}
\end{equation}
where \(\mathbb{I}[\cdot]\) is the indicator function, which equals \(1\)
when the enclosed condition is satisfied and \(0\) otherwise. \(S_k\) denotes the reliable geometric support of the current target
microbatch and \(\Delta_{\mathrm{scale}}^{(k)}\) the mean depth-scale
deviation between the post-update student and the fixed source-supervised
reference. During training, the student \(\Theta^{\mathrm{stu}}_{k}\) is updated through backpropagation, while the teacher \(\Theta^{\mathrm{tea}}_{k}\) is then updated by the gated exponential moving average (EMA):
\begin{equation}
    \Theta^{\mathrm{tea}}_{k}
    =
    \begin{cases}
        \mu\Theta^{\mathrm{tea}}_{k-1}
        +(1-\mu)\Theta^{\mathrm{stu}}_{k},
        & q_k=1,\\
        \Theta^{\mathrm{tea}}_{k-1},
        & q_k=0,
    \end{cases}
    \qquad k>K_{\mathrm{h}},
    \label{eq:teacher_schedule}
\end{equation}
where \(\mu\in[0,1)\) is the EMA decay coefficient. This update strategy
restricts target-domain changes in the student while preventing geometrically
unstable or scale-inconsistent student states from being propagated to the
teacher.

\subsection{Cross-View Supervision}
\label{subsec:crossview_supervision}

Since in-vivo video lacks geometric ground truth, we use cross-view
supervision to exploit the geometric relations between different views and
derive a learning signal from unlabeled target data. Using the teacher
predictions from Sec.~\ref{subsec:adaptation}, this signal forms
\(\mathcal{L}_{\mathrm{cv}}\) in Eq.~\eqref{eq:optimization}.

For each target sequence \(n\), we denote its frame index set by
\(\mathcal{S}\) and consider all directed view pairs
\(\mathcal{E}=\{(i,j)\mid i,j\in\mathcal{S},\,i\neq j\}\).
The cross-view objective is defined as
\begin{equation}
    \mathcal{L}_{\mathrm{cv}}
    =
    \frac{
        \sum_{n,(i,j),\mathbf{u}}
        \left[
            w_{n,ij}(\mathbf{u})
            \left|r_{n,ij}(\mathbf{u})\right|
        \right]
    }{
        \sum_{n,(i,j),\mathbf{u}}
        w_{n,ij}(\mathbf{u})
    },
    \label{eq:crossview}
\end{equation}
where \(r_{n,ij}(\mathbf{u})\) denotes the cross-view depth residual,
\(w_{n,ij}(\mathbf{u})\in[0,1]\) denotes its reliability weight which is
defined later, and
\(\mathbf{u}\) denotes a pixel location in source view \(i\). The summation covers
target sequences, directed view pairs, and valid projected pixels. 

For a transfer from view \(i\) to view \(j\), the residual compares the
student-predicted depth in view \(j\) with the teacher depth transferred from
view \(i\):
\begin{equation}
    r_{n,ij}(\mathbf{u})
    =
    \mathbf{D}^{\mathrm{stu}}_{n,j}
    \left(
        \mathbf{u}_{n,i\rightarrow j}(\mathbf{u})
    \right)
    -
    \widetilde{\mathbf{D}}^{\mathrm{tea}}_{n,i\rightarrow j}(\mathbf{u}),
    \label{eq:crossview_residual}
\end{equation}
where \(\mathbf{u}_{n,i\rightarrow j}(\mathbf{u})\) and
\(\widetilde{\mathbf{D}}^{\mathrm{tea}}_{n,i\rightarrow j}(\mathbf{u})\)
denote the projected location and transferred teacher depth in view \(j\),
respectively. They are obtained by unprojecting the teacher depth at
\(\mathbf{u}\) in view \(i\), transforming the resulting 3D point to view
\(j\), and projecting it into view \(j\):
\begin{equation}
\begin{aligned}
&\bigl(
\mathbf{u}_{n,i\rightarrow j}(\mathbf{u}),
\widetilde{\mathbf{D}}^{\mathrm{tea}}_{n,i\rightarrow j}(\mathbf{u})
\bigr) \\
&=
\Pi_j\!\left[
\mathbf{T}^{\mathrm{tea}}_{n,j}
\left(\mathbf{T}^{\mathrm{tea}}_{n,i}\right)^{-1}
\pi_i^{-1}\!\left(
\mathbf{u},
\mathbf{D}^{\mathrm{tea}}_{n,i}(\mathbf{u})
\right)
\right].
\end{aligned}
\label{eq:crossview_transfer}
\end{equation}
Here, \(\pi_i^{-1}\) denotes unprojection using the depth in view \(i\),
while \(\Pi_j\) denotes projection into view \(j\) together with extraction
of the corresponding depth.

\subsection{Hierarchical quasi-rigid reliability}
\label{subsec:hierarchical_reliability}

The cross-view supervision defined above relies on rigid geometric transfer
between views, which may not remain valid uniformly under non-rigid
colonoscopic observations. Motivated by the short-term and local deformation
priors used in NR-SLAM~\cite{rodriguez2024nr}, we consider that tissue motion
in temporally adjacent endoscopic observations can remain locally coherent
despite global non-rigidity. In particular, NR-SLAM assumes that camera motion
can dominate deformation over sufficiently short intervals and that small
tissue regions can exhibit approximately rigid local behavior. Rather than
explicitly estimating tissue deformation, we use these properties to determine
where rigid cross-view transfer remains sufficiently reliable for supervision.

Accordingly, we factor the cross-view reliability into three progressively
finer components:
\begin{equation}
    w_{n,ij}(\mathbf{u})
    =
    w^{\mathrm{seq}}_n\,
    w^{\mathrm{pair}}_{n,ij}\,
    w^{\mathrm{pix}}_{n,ij}(\mathbf{u}),
    \quad
    w_{n,ij}(\mathbf{u})\in[0,1],
    \label{eq:reliability}
\end{equation}
where \(w^{\mathrm{seq}}_n\), \(w^{\mathrm{pair}}_{n,ij}\), and \(w^{\mathrm{pix}}_{n,ij}(\mathbf{u})\) are the sequence-, pair-, and pixel-level reliability, respectively. We describe each level in turn below.

\subsubsection{Sequence-level Reliability}

The sequence-level term \(w^{\mathrm{seq}}_n\in\{0,1\}\) determines whether
a temporal window provides a reliable basis for target-domain supervision.
We define it as a binary validity gate:
\begin{equation}
    w^{\mathrm{seq}}_n
    =
    \mathbb{I}\!\left[
        C^{\mathrm{img}}_n
        \land
        C^{\mathrm{geo}}_n
        \land
        \rho_{\min}
        \leq
        \rho_n
        \leq
        \rho_{\max}
    \right],
    \label{eq:sequence_reliability}
\end{equation}
where \(C^{\mathrm{img}}_n\) denotes the image-validity condition for the
temporal window, including sufficient visibility and nondegenerate image
observations, while \(C^{\mathrm{geo}}_n\) denotes the geometric-validity
condition requiring finite teacher--student predictions and sufficient
cross-view support. The depth-ratio term \(\rho_n\) measures the relative
student--teacher depth scale, with \([\rho_{\min},\rho_{\max}]\) defining its
valid range. All validity thresholds are fixed throughout training. A temporal
window therefore contributes to target-domain supervision only when all three
conditions are satisfied.

\subsubsection{Pair-level Reliability}

Within a valid sequence, the pair-level reliability for a directed transfer
\(i\rightarrow j\) is defined as
\begin{equation}
    w^{\mathrm{pair}}_{n,ij}
    =
    h^{\mathrm{pair}}_{n,ij}
    \left[
        \beta
        +
        (1-\beta)
        \left(
            \rho^{\mathrm{sup}}_{n,ij}
            \rho^{\mathrm{mot}}_{n,ij}
            \rho^{\mathrm{geo}}_{n,ij}
        \right)^{1/3}
    \right],
    \label{eq:pair_reliability}
\end{equation}
where \(h^{\mathrm{pair}}_{n,ij}\in\{0,1\}\) rejects geometrically invalid
view pairs and \(\beta\in[0,1]\) sets the minimum contribution of a valid pair.
The normalized terms \(\rho^{\mathrm{sup}}_{n,ij}\),
\(\rho^{\mathrm{mot}}_{n,ij}\), and \(\rho^{\mathrm{geo}}_{n,ij}\) measure
cross-view support, relative camera motion, and transferred geometric
consistency, respectively. Specifically, they account for the spatial coverage
of valid transfers, depth-normalized translation and rotation, and depth
agreement with forward--backward reprojection consistency. The exponent
\(1/3\) forms their geometric mean.

\subsubsection{Pixel-level Reliability}

Even within a reliable view pair, individual correspondences may remain
unreliable due to local deformation, occlusion, or specular reflection.
We define the pixel-level reliability as
\begin{equation}
    w^{\mathrm{pix}}_{n,ij}(\mathbf{u})
    =
    m_{n,ij}(\mathbf{u})
    \left[
        \prod_{q\in\mathcal{C}_{\mathrm{pix}}}
        \eta^{q}_{n,ij}(\mathbf{u})
    \right]^{1/5},
    \label{eq:pixel_reliability}
\end{equation}
where
\(
\mathcal{C}_{\mathrm{pix}}
=
\{\mathrm{conf},\mathrm{dep},\mathrm{cyc},\mathrm{vis},\mathrm{app}\}
\)
indexes normalized local reliability scores.
The binary mask \(m_{n,ij}(\mathbf{u})\) rejects invalid or out-of-view
projections, nonpositive or nonfinite geometry, severe occlusion, and
specular regions. The terms
\(\eta^{\mathrm{conf}}\),
\(\eta^{\mathrm{dep}}\),
\(\eta^{\mathrm{cyc}}\),
\(\eta^{\mathrm{vis}}\), and
\(\eta^{\mathrm{app}}\)
measure teacher confidence, depth agreement, forward--backward consistency,
visibility, and appearance reliability, respectively. The exponent \(1/5\)
forms the geometric mean of these normalized scores.

Together, the sequence-, pair-, and pixel-level terms provide progressively
finer estimates of cross-view reliability. All reliability factors are computed from target-domain predictions without
target annotations and are treated as fixed during backpropagation.

    \setcounter{dbltopnumber}{2}
\renewcommand{\dbltopfraction}{0.95}
\renewcommand{\textfraction}{0.05}
\setlength{\dblfloatsep}{8pt}
\setlength{\dbltextfloatsep}{10pt}

\begin{figure*}[!t]
    \centering
    \includegraphics[width=0.98\textwidth]{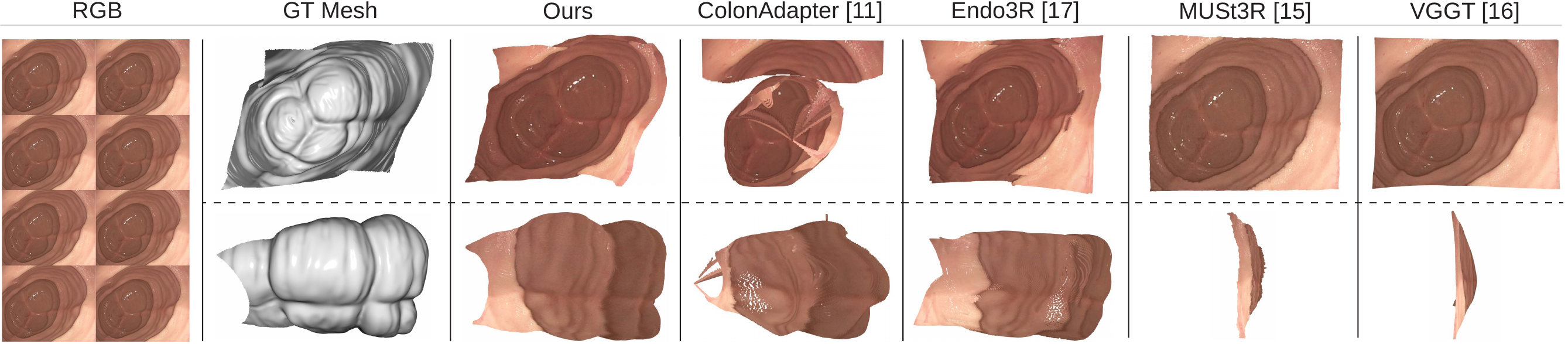}
    \caption{Qualitative comparison on representative C3VD cecum segment, showing the predicted pointmaps together with the ground-truth geometry.}
    \label{fig:c3vd_results}
\end{figure*}

\begin{table*}[!t] 
    \caption{
    Quantitative geometry estimation on SimCol3D and C3VD with fixed
    eight-frame input. Best results are in bold and second-best results are
    underlined among the external comparison methods.
    }
    \label{tab:c3vd_main} 
 
    \centering 
    \small 
    \setlength{\tabcolsep}{1.8pt} 
    \renewcommand{\arraystretch}{1.08} 
 
    \begin{tabularx}{\textwidth}{ 
        @{} 
        >{\raggedright\arraybackslash}m{1.95cm} 
        @{\hspace{2pt}} 
        *{12}{Y} 
        @{} 
    } 

        \specialrule{0.9pt}{0pt}{0pt}
 
        \multirow[c]{3}{*}[-1.7ex]{\textbf{Methods}} 
        & \multicolumn{6}{c}{\textbf{SimCol3D}~\cite{Rau2023}} 
        & \multicolumn{6}{c}{\textbf{C3VD}~\cite{Bobrow2023C3VD}} \\[1pt] 
 
        \cmidrule(l{8pt}r{8pt}){2-7} 
        \cmidrule(l{8pt}r{8pt}){8-13} 
 
        & \multicolumn{2}{c}{\textbf{Depth}} 
        & \multicolumn{2}{c}{\textbf{Pointmap}} 
        & \multicolumn{2}{c}{\textbf{Pose RPE}} 
        & \multicolumn{2}{c}{\textbf{Depth}} 
        & \multicolumn{2}{c}{\textbf{Pointmap}} 
        & \multicolumn{2}{c}{\textbf{Pose RPE}} \\[1pt] 
 
        \cmidrule(l{4pt}r{4pt}){2-3} 
        \cmidrule(l{4pt}r{4pt}){4-5} 
        \cmidrule(l{4pt}r{4pt}){6-7} 
        \cmidrule(l{4pt}r{4pt}){8-9} 
        \cmidrule(l{4pt}r{4pt}){10-11} 
        \cmidrule(l{4pt}r{4pt}){12-13} 
 
        & {\footnotesize\mbox{AbsRel $\downarrow$}} 
        & {\footnotesize\mbox{$\delta_1 \uparrow$}} 
        & {\footnotesize\mbox{L2 $\downarrow$}} 
        & {\footnotesize\mbox{p95 $\downarrow$}} 
        & {\footnotesize\mbox{Rot. $\downarrow$}} 
        & {\footnotesize\mbox{Trans. $\downarrow$}} 
        & {\footnotesize\mbox{AbsRel $\downarrow$}} 
        & {\footnotesize\mbox{$\delta_1 \uparrow$}} 
        & {\footnotesize\mbox{L2 $\downarrow$}} 
        & {\footnotesize\mbox{p95 $\downarrow$}} 
        & {\footnotesize\mbox{Rot. $\downarrow$}} 
        & {\footnotesize\mbox{Trans. $\downarrow$}} \\ 
 
        \hline 
 
        VGGT~\cite{Wang2025VGGT} 
        & \underline{0.2469} & \underline{0.6233} 
        & \underline{0.3513} & \underline{0.6720} 
        & 0.5390 & \underline{0.0174} 
        & 0.8542 & \underline{0.2861} 
        & \underline{0.5581} & \underline{0.8693} 
        & 0.2614 & 0.0039 \\ 
 
        MUSt3R~\cite{Cabon2025MUSt3R} 
        & 0.5425 & 0.3025 
        & 0.5583 & 0.9546 
        & 0.6728 & 0.0206 
        & 1.7013 & 0.0824 
        & 0.8547 & 1.3578 
        & 0.2141 & 0.0102 \\ 
 
        Endo3R~\cite{Guo2026Endo3R} 
        & 0.5989 & 0.0043 
        & 0.7207 & 1.3987 
        & 1.1159 & 0.0188 
        & \underline{0.5092} & 0.1008 
        & 0.7069 & 1.1973 
        & 0.6401 & 0.0066 \\ 
 
        {\footnotesize ColonAdap.}~\cite{Jiang2025ColonAdapter} 
        & 0.8409 & 0.0009 
        & 0.8499 & 1.5394 
        & \underline{0.5367} & 0.0197 
        & 0.8499 & 0.0039 
        & 0.8755 & 1.4171 
        & \textbf{0.0566} & \underline{0.0034} \\ 
 
        \rowcolor{gray!8} 
        \textbf{Ours} 
        & \textbf{0.1692} & \textbf{0.7738} 
        & \textbf{0.1309} & \textbf{0.2804} 
        & \textbf{0.4997} & \textbf{0.0102} 
        & \textbf{0.0675} & \textbf{0.9720} 
        & \textbf{0.0644} & \textbf{0.1449} 
        & \underline{0.1745} & \textbf{0.0032} \\ 
 
        \arrayrulecolor{gray!100}
        \specialrule{0.55pt}{2.2pt}{2.2pt}
        \arrayrulecolor{black}
 
        Ft VGGT$^\dagger$ 
        & 0.1436 & 0.8310 
        & 0.1340 & 0.3064 
        & 0.6204 & 0.0120 
        & 0.0608 & 0.9785 
        & 0.0562 & 0.1288 
        & 0.1608 & 0.0033 \\ 
 
        \specialrule{0.9pt}{0pt}{0pt}

    \end{tabularx} 
    
    \vspace{1mm}
    {\footnotesize
    $^\dagger$ Ft VGGT denotes the source-supervised fine-tuned VGGT and is
    reported separately as the source-only reference.
    \par}
\end{table*}

\section{Experiments}

\subsection{Datasets}
Labeled C3VD~\cite{Bobrow2023C3VD} phantom sequences and simulated
colonoscopy data from~\cite{zhang20213d} are used for
source-domain supervision. The unlabeled target domain comprises the
clinical in-vivo colonoscopy videos used in RNNSLAM~\cite{Ma2021RNNSLAM}.
C3VD provides registered camera poses, depth maps, calibrated camera models,
and surface geometry, while the simulated source data provide camera, depth,
and pointmap annotations. Training and evaluation sequences are disjoint.

Quantitative evaluation is performed on both phantom and synthetic data.
For C3VD, we use 113 fixed eight-frame windows from six held-out sequences:
\texttt{cecum\_t1\_b}, \texttt{cecum\_t2\_a},
\texttt{cecum\_t2\_c}, \texttt{cecum\_t4\_b},
\texttt{sigmoid\_t3\_b}, and \texttt{trans\_t2\_b}.
To additionally assess generalization to a fully synthetic environment, we
evaluate on SyntheticColon III from SimCol3D~\cite{Rau2023}, which is not used during
training. The evaluation set contains 225 fixed eight-frame windows from the
\texttt{O1}, \texttt{O2}, and \texttt{O3} trajectories, with 75 non-overlapping windows per trajectory. Evaluation uses the provided ground-truth camera poses and depth maps, with metric errors reported in centimetres.

For the clinical target domain, evaluation uses 21 fixed eight-frame
RGB-only windows from the in-vivo videos~\cite{Ma2021RNNSLAM}. Because metric geometric ground
truth is unavailable for these sequences, the in-vivo results characterize
geometric plausibility and cross-view behavior rather than absolute
reconstruction accuracy.

\subsection{Experimental Setup}

\subsubsection{Implementation details.}

Our method is initialized from the source-supervised VGGT~\cite{Wang2025VGGT} checkpoint obtained
after 4,000 updates and adapted for a further 6,000 updates. Each update uses
labeled source and unlabeled target microbatches at a \(3{:}1\) ratio.
Rank-\(16\) LoRA adapters are applied to the final 12 global-attention blocks,
with the remaining backbone frozen. We use AdamW with cosine
learning-rate decay and a peak LoRA learning rate of \(2\times10^{-5}\).

The teacher is fixed for the first \(K_{\mathrm{h}}=500\) updates and then
updated by the stability-gated EMA in Eq.~\eqref{eq:teacher_schedule} with \(\mu=0.996\).
The target weight \(\lambda_t(k)\) increases from \(0.025\) to \(0.5\) during
warm-up, with \(\lambda_{\mathrm{anc}}=0.05\) and \(\beta=0.1\).
All reliability thresholds are fixed across training and experiments. Teacher and student use the
same eight-frame target windows, yielding 56 directed view pairs. Training is
performed on two NVIDIA L40 GPUs.

\subsubsection{Baseline and Evaluation protocol.}
All methods receive the same fixed eight-frame RGB observations at inference.
We compare our method with the pretrained VGGT~\cite{Wang2025VGGT},
source-supervised fine-tuned VGGT,
MUSt3R~\cite{Cabon2025MUSt3R}, Endo3R~\cite{Guo2026Endo3R}, and
ColonAdapter~\cite{Jiang2025ColonAdapter}.

\noindent\textbf{Calibration and metrics.}
C3VD uses the provided Scaramuzza calibration, whereas
SimCol3D~\cite{Rau2023} uses its native pinhole model. Both follow the
camera-axis depth convention. Depth is evaluated over valid pixels without
post-hoc scale alignment using AbsRel and \(\delta_1\) (threshold \(1.25\)).
Direct pointmaps are evaluated in the normalized first-camera frame using mean
and 95th-percentile Euclidean errors (L2 and p95), while camera pose is
evaluated using rotational and translational RPE over consecutive-frame
relative transforms. All metrics are averaged over the fixed evaluation
windows.

\begin{figure*}[!t]
    \centering
    \includegraphics[width=\textwidth]{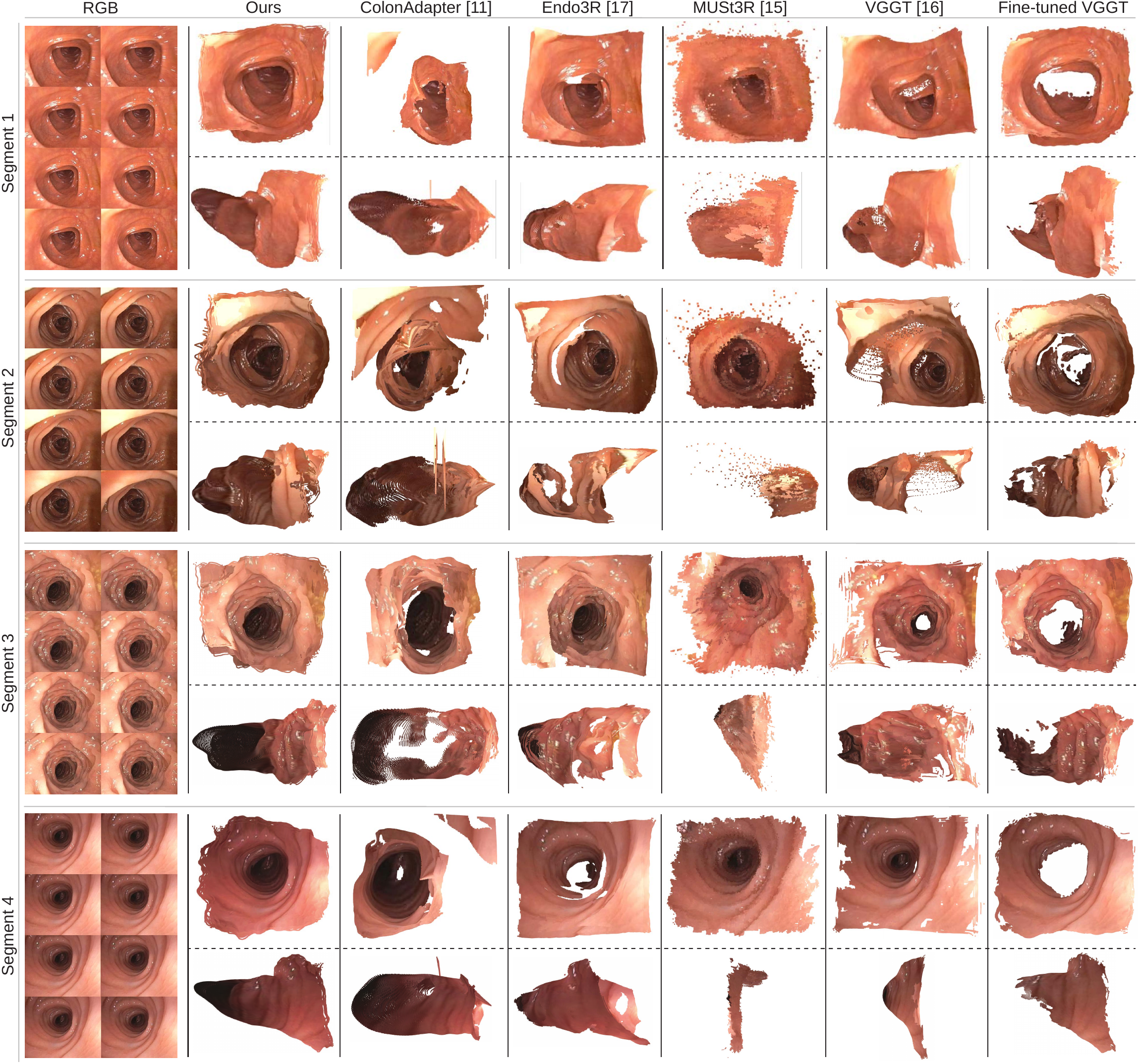}
    \caption{Qualitative comparison on representative in-vivo sequences,
    showing the reconstructed geometry of different methods.}
    \label{fig:invivo_results}
\end{figure*}

\noindent\textbf{Baseline handling.}
MUSt3R and Endo3R directly predict pointmaps and derive depth and pose from
them. ColonAdapter follows its official centered-pinhole global-alignment
pipeline. Its end-to-end outputs are evaluated using the same input windows
and metrics, while its internal camera parameterization differs from the
native benchmark calibration.

\noindent\textbf{In-vivo evaluation.}
For the in-vivo data~\cite{Ma2021RNNSLAM}, where metric geometric ground truth
is unavailable, we report descriptive statistics of the predicted depth,
point geometry, and pose.

\subsection{Evaluation on SimCol3D and C3VD}

We evaluate coupled geometry estimation on SimCol3D and the held-out C3VD
sequences. Table~\ref{tab:c3vd_main} compares our method with VGGT, MUSt3R,
Endo3R, and ColonAdapter as external methods, while fine-tuned VGGT is separately reported in the table because it serves as a source-only
reference for evaluating geometry preservation after target-domain adaptation,
rather than as an external comparison method. Accordingly, the best and second-best results
are determined only among the external comparison methods.

Among the external methods, our method ranks first in 91.7\% of the reported
metrics across depth, pointmap, and camera-pose estimation. Relative to the
original VGGT, our method reduces Abs Rel and direct pointmap L2 error by
31.5\% and 62.7\% on SimCol3D, and by 92.1\% and 88.5\% on C3VD,
respectively. The improvements are particularly clear in dense geometry, with
substantially lower mean and tail pointmap errors while maintaining competitive
camera-pose accuracy. Compared with ColonAdapter, our method performs better on
all SimCol3D metrics and five of the six C3VD metrics, with ColonAdapter only
achieving lower rotational RPE on C3VD.

Fig.~\ref{fig:c3vd_results} provides complementary qualitative evidence.
Our method reconstructs more complete and coherent tubular geometry, with less
scale distortion and local collapse than the competing methods, consistent with
the quantitative improvements in Table~\ref{tab:c3vd_main}.

Compared with fine-tuned VGGT, our adapted model remains close on the controlled-domain
metrics while improving the SimCol3D pointmap and pose results and the C3VD
translational RPE. However, strong performance on these controlled benchmarks
does not necessarily imply robust generalization to the more complex in-vivo
domain, as examined in the following subsection.

These
results indicate that the proposed adaptation largely preserves the
source-supervised geometric performance after introducing unlabeled
target-domain supervision. This preservation is supported by continued source
supervision, which anchors the learned coupled geometry, together with
restricted target updates and reliability-weighted cross-view supervision that
limit the influence of unreliable target-domain gradients.
Although our adapted model shows slight decreases relative to fine-tuned VGGT on
several metrics, it produces more complete and geometrically
consistent reconstructions on unseen in-vivo sequences, as demonstrated in the
following subsection.

\begin{table}[t]
    \caption{Ablation study of the adaptation progression.}
    \label{tab:adaptation_ablation}
    \centering
    \small
    \setlength{\tabcolsep}{1.5pt}
    \renewcommand{\arraystretch}{1.05}
    \begin{tabular}{@{}l@{\hspace{2pt}}
        >{\centering\arraybackslash}p{39pt}@{\hspace{3pt}}
        >{\centering\arraybackslash}p{39pt}@{\hspace{3pt}}
        >{\centering\arraybackslash}p{39pt}@{\hspace{3pt}}
        >{\centering\arraybackslash}p{39pt}@{}}
        \specialrule{0.9pt}{0pt}{0pt}
        Variant & In-vivo & Depth & Pointmap & Pose RPE \\[1pt]
        & CV$^{*}$ $\downarrow$ & Rel $\downarrow$ &
        L2 $\downarrow$ & Rot. $\downarrow$ \\
        \hline

        Fine-tuned VGGT
        & 0.0764 & 0.0608 & 0.0562 & 0.1608 \\

        {\footnotesize \quad + EMA}
        & 0.0435 & 0.0862 & 0.0915 & 0.2217 \\

        {\footnotesize \quad + Gated EMA}
        & 0.0520 & 0.0728 & 0.0736 & 0.1825 \\

        {\footnotesize \quad + Sequence-level}
        & 0.0508 & 0.0717 & 0.0711 & 0.1829 \\

        {\footnotesize \quad + Pair-level}
        & 0.0489 & 0.0693 & 0.0676 & 0.1788 \\

        {\footnotesize \quad + Pixel-level (Ours)}
        & 0.0479 & 0.0675 & 0.0644 & 0.1745 \\

        \specialrule{0.9pt}{0pt}{0pt}
    \end{tabular}

    \vspace{1mm}
    {\footnotesize
    $^{*}$ In-vivo CV denotes the unweighted cross-view depth residual used to evaluate
    geometric consistency.
    \par}
\end{table}

\subsection{Evaluation on In-Vivo Dataset}
\label{sec:invivo}

C3VD and SimCol3D provide quantitative evaluation with geometric ground truth,
however, such annotations are unavailable for real in-vivo colonoscopy. To
further assess target-domain transfer, we compare our method with existing
approaches on real in-vivo video~\cite{Ma2021RNNSLAM}.
Fig.~\ref{fig:invivo_results} shows four representative cases with different
lumen appearances and geometric conditions.

In Segment~1, close-range observations and strong specular regions lead
ColonAdapter and MUSt3R to produce more fragmented or noisy geometry, while
VGGT and fine-tuned VGGT recover a shorter tubular extent. Our method preserves a more
continuous lumen surface, suggesting that pixel-level reliability helps reduce
the influence of locally unreliable supervision.

Segment~2 contains repetitive folds and varying view overlap. Here, VGGT and
MUSt3R exhibit substantial incompleteness or collapse, while ColonAdapter and
Endo3R recover shorter or less coherent tubular structures. In contrast, our
method maintains a substantially longer and more complete reconstruction,
indicating greater robustness to weakly supported cross-view transfers through
pair-level reliability.

In Segment~3, local appearance and tissue changes result in larger missing
regions or distorted geometry for several competing methods, whereas our method
maintains a more complete asymmetric lumen opening and surrounding wall
structure. This reduced local degradation aligns with the intended role of
pixel-level reliability in limiting unreliable spatial correspondences.

In Segment~4, the nearly axial view and weak texture provide limited geometric
evidence. VGGT, MUSt3R, Endo3R, and fine-tuned VGGT recover only limited longitudinal
structure, whereas our method retains a more complete tubular extent. This
comparison supports the role of gated teacher updates and source-preserving
adaptation in maintaining stable geometry when target-domain supervision is
weak.

Overall, the comparison shows that the advantage of our method is not confined
to a single imaging condition. Compared with general geometry models,
source-only fine-tuned VGGT, and endoscopy-specific alternatives, our method exhibits
less fragmentation, local collapse, and truncation across different in-vivo
failure modes. These differences are consistent with reliability-aware
cross-view supervision reducing unreliable target constraints while
source-preserving adaptation maintains the learned geometric prior.

The contrast with our method further shows that the strong performance of
fine-tuned VGGT on the controlled benchmarks mainly reflects effective learning
within the labeled source distribution, rather than robust transfer to the
clinical domain. Its clear degradation on in-vivo sequences indicates
that source-only fine-tuning is insufficient to bridge the source--target
domain gap. These results support the need for target-domain adaptation with
reliability-aware cross-view supervision while preserving the learned source
geometry.

\subsection{Ablation Study}

The results in Table~\ref{tab:c3vd_main} and Fig.~\ref{fig:invivo_results}
demonstrate the advantage of our method in quantitative geometry estimation on
controlled benchmarks and qualitative reconstruction on real in-vivo
colonoscopy. In this ablation study, we further investigate the effectiveness
of its individual components, including EMA-based target adaptation, gated
teacher updates, and the sequence-, pair-, and pixel-level reliability.

Table~\ref{tab:adaptation_ablation} shows that the full model achieves the best
overall balance between in-vivo cross-view consistency and retention of the
source-supervised geometry. All variants follow the same evaluation protocol,
using an unweighted cross-view residual for in-vivo consistency and C3VD depth,
pointmap, and pose errors for source-domain retention. Although uniform EMA
achieves the lowest in-vivo residual, it substantially degrades the C3VD
metrics. Conversely, fine-tuned VGGT retains stronger controlled-domain accuracy but
does not exploit target-domain supervision. The full model achieves a low in-vivo cross-view residual while substantially
reducing the C3VD degradation observed with uniform EMA, indicating more
effective target adaptation without sacrificing as much source-domain geometry.

The progressive comparison further demonstrates the contribution of each
component. Gated teacher updates recover much of the source-domain accuracy
lost with uniform EMA while retaining improved in-vivo consistency.
Sequence-level reliability further improves the cross-view, depth, and
pointmap results, while pair-level reliability improves all four metrics by
reducing weakly supported view transfers. Pixel-level reliability provides
additional gains by suppressing unreliable local correspondences.

Overall, the ablation confirms that gated teacher updates and hierarchical
reliability progressively strengthen target-domain adaptation while limiting
degradation of the source-supervised geometry, yielding the best
adaptation--preservation balance in the complete model.

\section{Conclusion and Limitations}

To address the challenges of adapting coupled geometric prediction to
unlabeled in-vivo colonoscopy, where domain shift and non-rigid tissue motion
limit the reliability of cross-view supervision, we presented Colon3R, a cross-domain framework for adapting a visual
geometry foundation model from labeled phantom and simulated colonoscopy data
to unlabeled in-vivo video. Our method adapts the source-supervised geometry prior to unlabeled in-vivo video
through cross-view supervision guided by hierarchical quasi-rigid reliability,
while source-preserving updates retain the coupled geometry learned from
labeled data. Experiments on labeled C3VD and SimCol3D demonstrate superior overall
performance over state-of-the-art methods in coupled geometry estimation,
while real in-vivo comparisons show substantially more complete and
geometrically consistent reconstructions under clinical domain shift.

While our method handles non-rigid observations by identifying reliable
cross-view constraints, it does not explicitly estimate tissue deformation.
Future work will investigate integrating deformable SLAM to jointly model
camera motion and tissue deformation, which may further improve geometric
consistency under larger and more complex non-rigid motion.

\ifhasreferences
    \bibliographystyle{IEEEtran}
    \bibliography{refs}
\fi

\end{document}